\documentclass{article}

\PassOptionsToPackage{numbers,sort&compress,comma}{natbib}
\usepackage{arxiv}

\usepackage[utf8]{inputenc}
\usepackage[T1]{fontenc}
\usepackage{hyperref}
\usepackage{url}
\usepackage{booktabs}
\usepackage{amsfonts}
\usepackage{amssymb}
\usepackage{nicefrac}
\usepackage{microtype}
\usepackage{graphicx}
\usepackage{doi}
\usepackage{amsmath}
\usepackage{xcolor}
\usepackage{subcaption}
\usepackage{url}
\usepackage{multirow}

\title{
Improving Imbalanced Multi-Label Chest X-Ray Diagnosis via CBAM-Enhanced CNN Backbones
\thanks{An earlier version of this work was accepted at the FPT International Conference on Emerging Trends in Computing (FETC).}
}

\author{
\href{mailto:duynhse183995@fpt.edu.vn}{Nguyen Huu Duy} \\
Department of Artificial Intelligence\\
FPT University\\
Ho Chi Minh City, Vietnam\\
\texttt{duynhse183995@fpt.edu.vn}
\And
\href{mailto:duyhkse184883@fpt.edu.vn}{Hoang Khuong Duy} \\
Department of Artificial Intelligence\\
FPT University\\
Ho Chi Minh City, Vietnam\\
\texttt{duyhkse184883@fpt.edu.vn}
\And
\href{mailto:nguhcv@fe.edu.vn}{Huynh Cong Viet Ngu\thanks{Correspondence can be addressed to Huynh Cong Viet Ngu at \texttt{nguhcv@fe.edu.vn}.}} \\
Department of Computing Fundamental\\
FPT University\\
Ho Chi Minh City, Vietnam\\
\texttt{nguhcv@fe.edu.vn}
}

\renewcommand{\shorttitle}{\textit{arXiv} Template}

\hypersetup{
pdftitle={A template for the arxiv style},
pdfsubject={q-bio.NC, q-bio.QM},
pdfauthor={David S.~Hippocampus, Elias D.~Striatum},
pdfkeywords={First keyword, Second keyword, More},
}

\begin{document}
\maketitle

\begin{abstract}
Chest radiography is a widely used imaging modality for thoracic disease diagnosis, yet its conventional interpretation remains time-consuming and heavily dependent on expert knowledge. While deep learning has improved diagnostic efficiency through automated feature extraction, challenges such as class imbalance and the localization of multiple co-existing pathologies remain unsolved. In this paper, inspired by the strength of Convolutional Block Attention Module (CBAM) in feature refinement and the capability of CNN blocks in feature extraction, we propose a strategy to integrate CBAM into traditional CNN blocks to enhance performance in multi-label classification tasks. Our method achieves a mean AUC of 0.8695 on ChestXray14 dataset, outperforming several state-of-the-art baselines.Our source code is available at: \url{https://github.com/NNNguyenDuyyy/FETC_CBAM_Enhanced_CNN.git}
\end{abstract}


\section{Introduction}
Chest X-rays (CXRs) are among the most accessible and widely utilized imaging modalities for the screening and diagnosis of thoracic diseases such as Pneumonia, Edema, Fibrosis, and Cardiomegaly. However, manual interpretation of CXRs is time-consuming and requires expert radiological knowledge, motivating the adoption of Deep Learning (DL) techniques for automated clinical decision support. A major challenge in this domain is the \textit{multi-label classification} problem, where a single radiograph may contain multiple co-existing pathological findings. This complexity is further exacerbated by \textit{severe class imbalance}, as frequently occurring conditions dominate large-scale datasets such as ChestXray14~\cite{1}, MIMIC-CXR~\cite{2}, and CheXpert~\cite{3}, resulting in biased learning behavior and degraded performance on rare but clinically important diseases.

Convolutional Neural Networks (CNNs) have driven significant progress in automated CXR analysis due to their strong visual feature extraction capabilities~\cite{4,5,6,7}. These models learn complex spatial representations from large annotated datasets and have achieved diagnostic accuracy close to expert radiologists. Nonetheless, standard CNNs suffer from limited receptive fields, reducing their ability to capture long-range dependencies—crucial for identifying spatially distant yet semantically correlated abnormalities.

To overcome this limitation, Transformer-based architectures have emerged as effective alternatives. For instance, Swinchex~\cite{8} exploits the hierarchical structure of Swin Transformers~\cite{9}, while MedViT~\cite{10} integrates domain-specific inductive bias into a ViT~\cite{11} backbone. Despite their success, such models require substantial training data and computational resources, making them less feasible in real-world clinical scenarios.

As a trade-off between modeling capacity and efficiency, lightweight attention-based CNNs have been proposed. These models integrate attention mechanisms—such as channel or spatial attention—into CNN backbones to emphasize diagnostically relevant regions while maintaining computational tractability. For example, \cite{12} introduces a channel attention module into a DenseNet~\cite{13} backbone to enhance feature selection, while \cite{14} incorporates spatial attention via the fusion of local and global features in a VGG16~\cite{15}-based architecture. However, these designs typically focus on either channel-wise or spatial-wise attention in isolation, limiting their ability to capture comprehensive contextual information. Moreover, the issue of class imbalance remains inadequately addressed, adversely affecting model robustness and generalization.

In this paper, we propose a practical integration strategy that enhances standard CNN backbones with attention-based mechanisms to improve multi-label chest X-ray classification. Our main contributions are summarized as follows:

\begin{itemize}
    \item We present a CBAM-Enhanced CNN framework that incorporates the Convolutional Block Attention Module (CBAM)~\cite{16} into widely adopted backbones such as DenseNet121 and VGG16. Rather than designing new attention modules, we focus on an effective integration strategy to refine feature representations for multi-label classification without requiring pixel-level annotations.

    \item We introduce a principled module placement scheme to identify optimal insertion points for CBAM within CNN architectures. This scheme balances the backbone’s low-level feature extraction capabilities with high-level semantic refinement, improving sensitivity to subtle pathological patterns.

    \item We conduct an ablation study investigating the effects of Focal Loss~\cite{17}, one-stage training, and a curriculum-inspired strategy adapted from~\cite{7}. This analysis sheds light on how different training configurations affect performance under severe class imbalance, especially for the dominant “No-Finding” label and rare disease categories.

    \item We evaluate our approach on the ChestXray14 dataset, achieving a mean AUC of \textbf{0.8695}, surpassing several state-of-the-art methods~\cite{1,5,7}, particularly in detecting underrepresented conditions.
\end{itemize}

The remainder of this paper is organized as follows: Section~2 reviews related work on chest X-ray classification and attention-augmented CNNs. Section~3 presents our proposed method. Section~4 details experimental results and comparisons. Section~5 concludes the paper.

\section{Related Work}

\subsection{Deep Learning Techniques for Multi-Label Chest X-ray Classification}

\begin{figure}[b]
\centerline{\includegraphics[width=12cm]{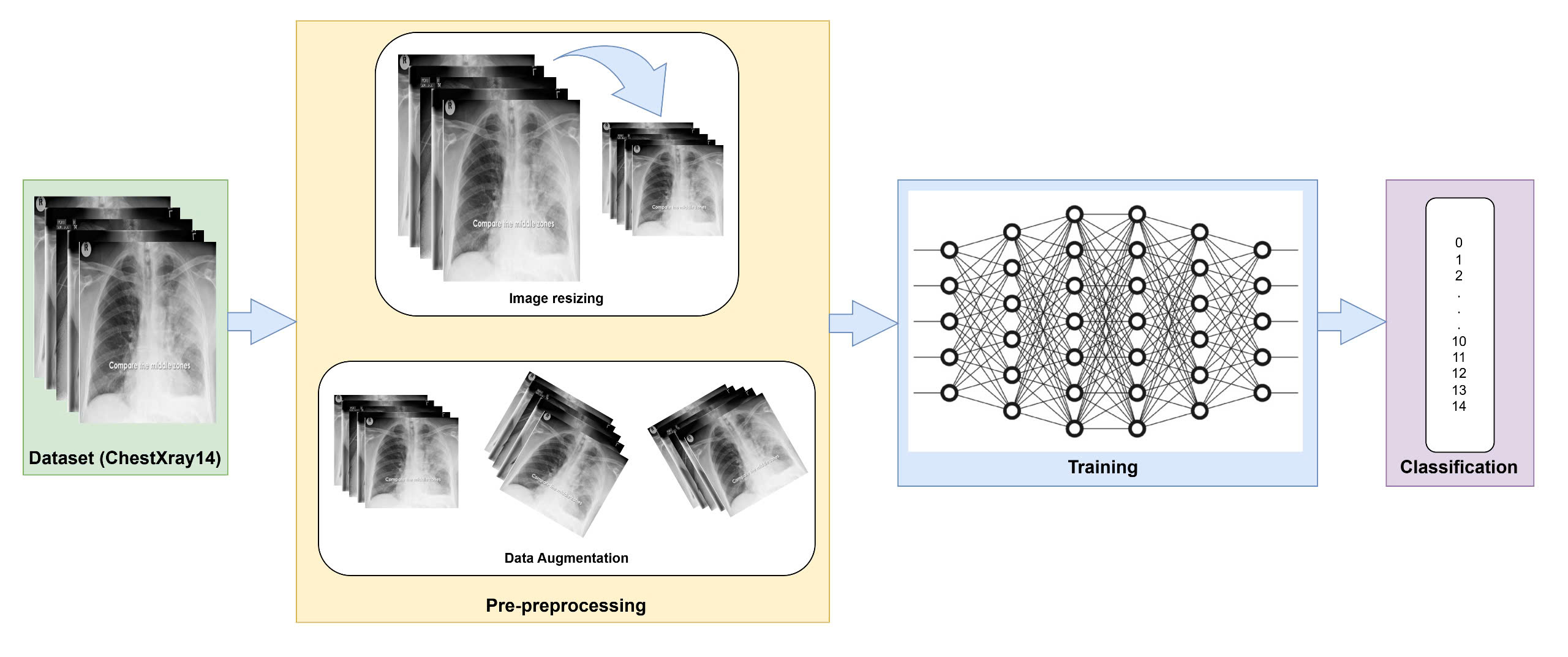}}
\vspace*{8pt}
\caption{The workflow of Learning Technique for ChestXray14 multi-label classification.}
\label{dl_based_learning}
\end{figure}

Most existing approaches to multi-label classification on the ChestXray14 dataset follow a common deep learning pipeline. As illustrated in Fig.~\ref{dl_based_learning}, the workflow typically begins with preprocessing steps such as image resizing and data augmentation to normalize input dimensions and enhance data diversity. The resulting chest X-ray images are then processed by deep neural networks (DNNs), including Convolutional Neural Networks (CNNs), Transformer-based architectures, or hybrid CNN-attention models. Finally, high-level features extracted by these networks are passed through a linear classification head to generate multi-label predictions corresponding to thoracic pathologies.

\subsection{CBAM: Convolutional Block Attention Module}

\begin{figure}[h]
\centerline{\includegraphics[width=8cm]{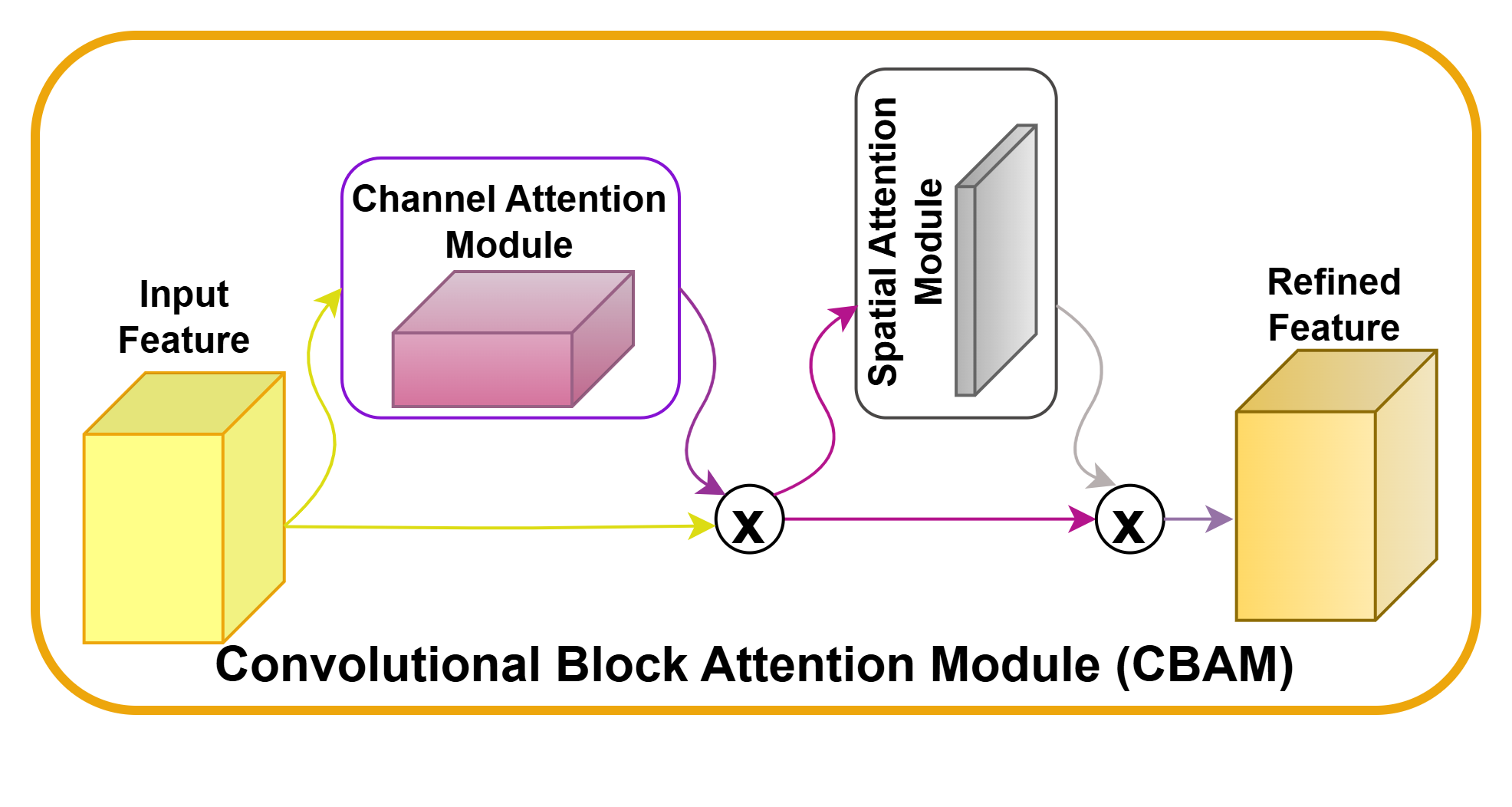}}
\vspace*{8pt}
\caption{The overview of CBAM.}
\label{CBAM}
\end{figure}

The Convolutional Block Attention Module (CBAM)~\cite{16} is a lightweight attention mechanism designed to enhance feature representations in Convolutional Neural Networks (CNNs). As illustrated in Fig.~\ref{CBAM}, CBAM applies attention sequentially along the channel and spatial dimensions by generating distinct attention maps. The \textit{channel attention module} leverages both global average pooling and max pooling, followed by shared multi-layer perceptrons, to capture inter-channel dependencies. Subsequently, the \textit{spatial attention module} aggregates these refined features to emphasize salient spatial regions. Owing to its modularity and efficiency, CBAM has been widely integrated into various deep learning architectures for tasks including image classification, object detection, and medical image analysis.

\section{Proposed Method}
\subsection{Overall Architecture}

\begin{figure}[h]
\centerline{\includegraphics[width=12cm]{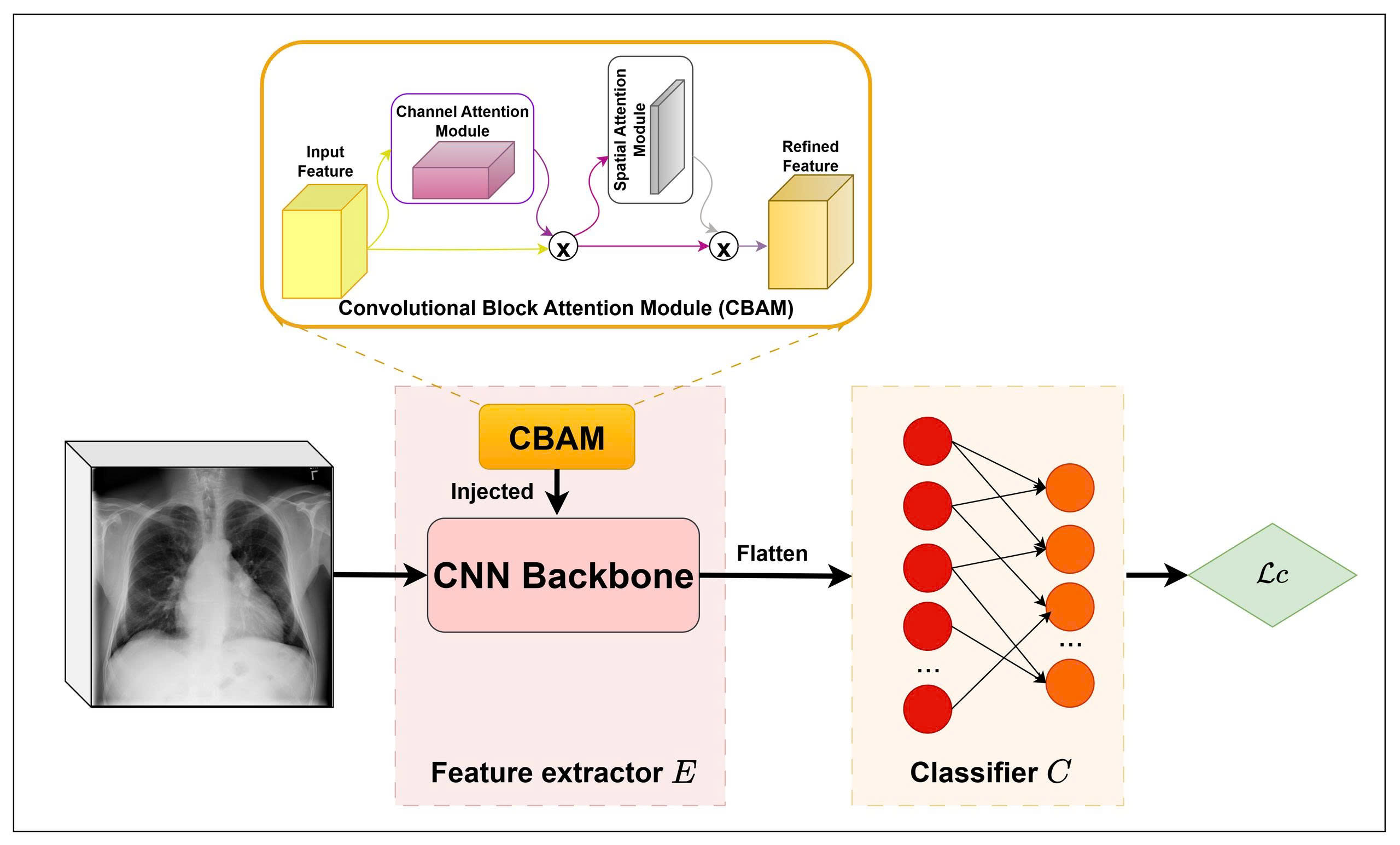}}
\vspace*{8pt}
\caption{Overview of CBAM-Enhanced CNN Backbones.}
\label{overall_architecture}
\end{figure}

We propose a lightweight modification to standard CNN backbones by incorporating an attention mechanism inspired by the Convolutional Block Attention Module (CBAM)~\cite{16}, aiming to emphasize discriminative regions relevant to the classification task. The design combines spatial and channel attention in an efficient manner, enabling the model to identify task-relevant features using only image-level supervision. To ensure effective integration, the attention module is strategically embedded within selected convolutional blocks, preserving low-level feature extraction while enhancing high-level semantic representations. The CBAM-enhanced feature maps are subsequently passed through a linear classifier \(C\). The training procedure follows the two-stage learning paradigm introduced in~\cite{7}. An overview of the architecture is shown in Fig.~\ref{overall_architecture}.

\subsection{CBAM-Enhanced for Feature extractor \(E\)}

\subsubsection{Construction of CBAM-Enhanced CNN Blocks}

To refine high-level features and guide the network's focus toward informative regions, we integrate Convolutional Block Attention Module (CBAM)~\cite{16} into selected stages of the CNN backbone. At each designated stage, the original convolutional block is augmented with a CBAM module as follows:
\begin{equation}
x^{(k)} = 
\begin{cases} 
B_k(x^{(k-1)}), & k \notin S, \\
A_k(B_k(x^{(k-1)})), & k \in S,
\end{cases}
\end{equation}
where \( B_k(\cdot) \) denotes the \( k \)-th convolutional block, \( A_k(\cdot) \) represents the attention module applied at stage \( k \), and \( S \) indicates the set of stages with attention integration. The attention maps generated by CBAM act as soft feature masks, amplifying task-relevant regions while suppressing irrelevant background information, thereby enhancing the quality of intermediate feature representations.

\subsubsection{Strategic CBAM-Enhanced CNN Blocks Placement for Feature Extractor \(E\)}

\begin{figure}[h]
\centerline{\includegraphics[width=12cm]{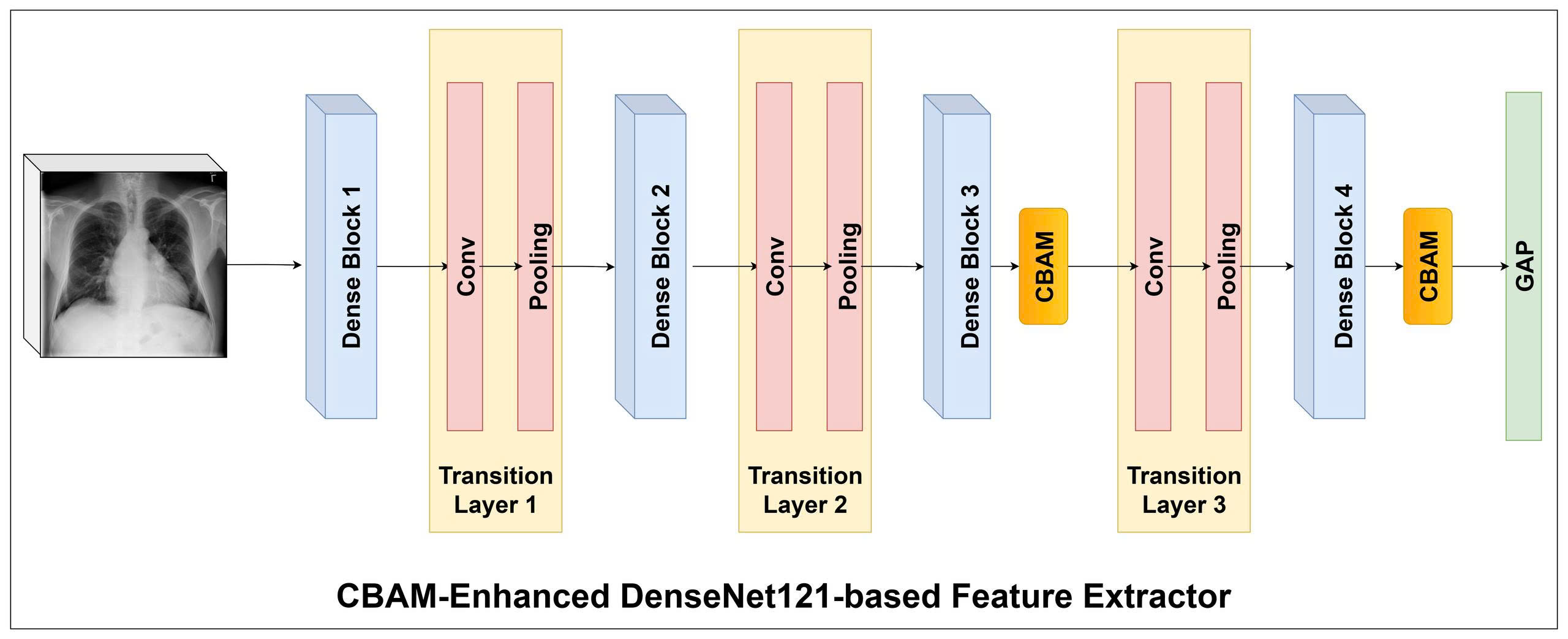}}
\vspace*{8pt}
\caption{CBAM-Enhanced DenseNet121-based Feature Extractor.}
\label{dense_extractor}
\end{figure}

We explore two variants of the feature extractor \(E\), based on DenseNet121 and VGG16 backbones. For the DenseNet121 architecture, Convolutional Block Attention Modules (CBAM)~\cite{16} are inserted after the third and fourth dense blocks, as illustrated in Fig.~\ref{dense_extractor}. These stages correspond to deeper semantic layers, where high-level abstractions are formed, making them suitable for enhancing discriminative attention. The dense connectivity pattern of DenseNet promotes efficient information flow across layers, and the selective integration of CBAM further strengthens spatial and channel-wise feature refinement critical to multi-label classification.

\begin{figure}[h]
\centerline{\includegraphics[width=12cm]{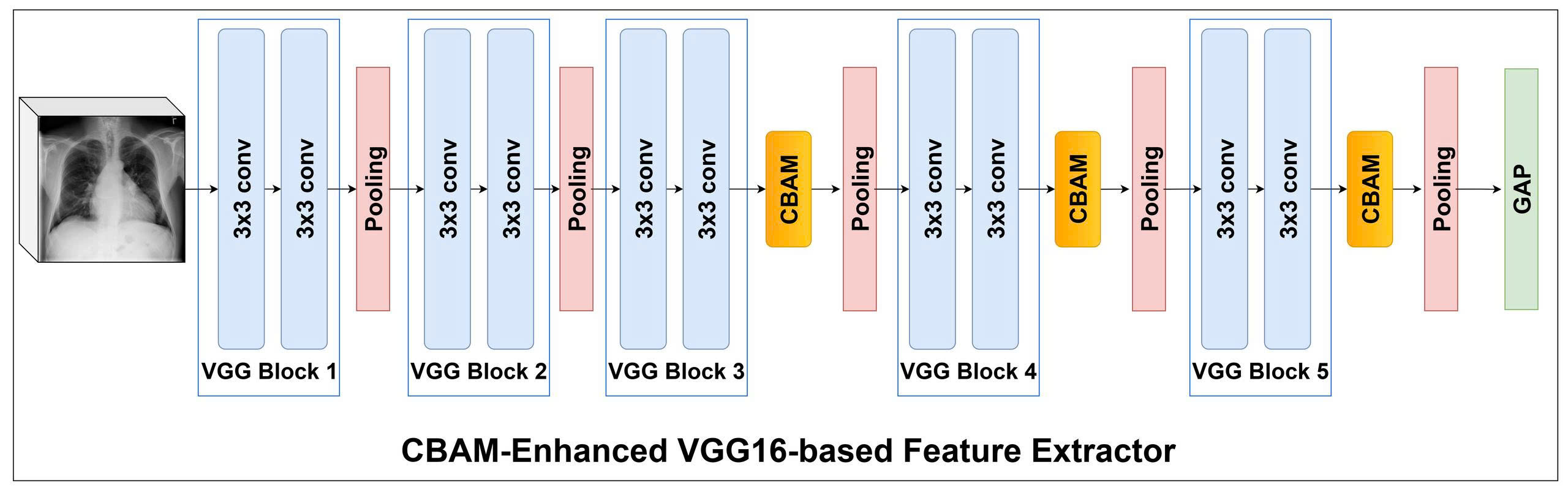}}
\vspace*{8pt}
\caption{CBAM-Enhanced VGG16-based Feature Extractor.}
\label{vgg_extractor}
\end{figure}

For the VGG16-based feature extractor, Convolutional Block Attention Modules (CBAM)~\cite{16} are inserted after blocks 3, 4, and 5, as illustrated in Fig.~\ref{vgg_extractor}. This configuration leverages the sequential and hierarchically deep structure of VGG16 to enable progressive refinement of attention across increasing semantic levels.

By integrating attention modules at deeper stages of the backbone, the model preserves fine-grained spatial details in early layers while enhancing high-level semantic representations where contextual understanding is critical. Notably, this refinement is achieved using only image-level supervision, without relying on explicit localization annotations.

\section{Results And Discussion}

\subsection{Dataset and Preprocessing}

We evaluate our method on the ChestXray14 dataset~\cite{1}, a large-scale benchmark comprising 112{,}120 frontal-view chest X-ray images labeled with 14 thoracic disease categories. As a standard benchmark for multi-label classification in medical imaging, ChestXray14 serves as a widely-used dataset for evaluating disease detection models. To ensure clinically valid evaluation and avoid potential information leakage, we follow the patient-wise split and preprocessing protocol proposed in~\cite{7}.

\subsection{Performance Comparison with State-of-the-Art Methods}

As summarized in Table~\ref{compare_SOTA}, our proposed CBAM-enhanced feature extractors based on DenseNet121 and VGG16 achieve state-of-the-art mean AUCs of \textbf{0.8681} and \textbf{0.8695}, respectively, on the ChestXray14 dataset. These results substantially outperform previous benchmarks, including foundational baselines such as Wang et al.~\cite{1} (0.7451), CheXNet~\cite{5} (0.841), and more recent approaches like SynthEnsemble~\cite{7} (0.854).

Beyond overall performance, our models exhibit strong discriminative capability on rare yet clinically significant thoracic conditions, demonstrating resilience to class imbalance. For \textit{Pneumonia}, the CBAM-DenseNet121 achieves an AUC of \textbf{0.7791}, while the CBAM-VGG16 yields a higher AUC of \textbf{0.7915}. In the case of \textit{Hernia}, one of the rarest pathologies in ChestXray14, the CBAM-VGG16 model attains an AUC of \textbf{0.9529}, while CBAM-DenseNet121 achieves a new peak performance of \textbf{0.9660}. Similarly, for \textit{Emphysema}, the models achieve AUCs of \textbf{0.9426} (DenseNet121) and \textbf{0.9422} (VGG16), both exceeding prior results. For \textit{Edema}, the CBAM-DenseNet121 and CBAM-VGG16 models obtain AUCs of \textbf{0.9214} and \textbf{0.9216}, respectively. These findings highlight the effectiveness of our proposed architecture not only in improving aggregate metrics, but also in enhancing detection of subtle and low-prevalence pathologies—an essential attribute for practical deployment in clinical environments.

\begin{table}
\caption{Comparison of AUC scores with state-of-the-art methods on ChestXray14 dataset.\label{compare_SOTA}}

{\begin{tabular}{@{}cccccc@{}} \toprule
Pathology & Wang et al.\cite{1} & Rajpurkar et al.\cite{5} & Ashraf et al.\cite{7}  & Ours & Ours\\
&&&& DenseNet121 & VGG16 \\\midrule
Atelectasis & 0.7003 & 0.8094 & 0.83390 & \textit{0.8396} & \textbf{0.8509}\\
Cardiomegaly & 0.8100 & 0.9248 & 0.91954 & \textit{0.9382} & \textbf{0.9386} \\
Effusion & 0.7585 & 0.8638 & 0.88977 & \textit{0.8994} & \textbf{0.9014}\\
Infiltration & 0.6614 & 0.7345 & \textit{0.74102} & 0.7387 & \textbf{0.7444} \\
Mass & 0.6933 & 0.8676 & 0.87315  & \textbf{0.8879} & \textit{0.8860}\\
Nodule & 0.6687 & 0.7802 & 0.80611 & \textit{0.8153} & \textbf{0.8315}\\
Pneumonia & 0.6580 & 0.7680 & 0.77648 & \textit{0.7791} & \textbf{0.7915} \\
Pneumothorax & 0.7993 & 0.8887 & 0.90164& \textit{0.9169} & \textbf{0.9189} \\
Consolidation & 0.7032 & 3 & 0.81575 & \textbf{0.8305} & \textit{0.8242} \\
Edema & 0.8052 & 0.8878 & 0.91034 & \textit{0.9214} & \textbf{0.9216} \\
Emphysema & 0.8330 & 0.9371 & 0.92946 & \textbf{0.9426} & \textit{0.9422} \\
Fibrosis & 0.7859 & 0.8047 & 0.83347 & \textbf{0.8417} & \textit{0.8384}\\
Pleural Thickening & 0.6835 & 0.8062 & 0.81270 & \textbf{0.8362} & \textit{0.8311} \\
Hernia & 0.8717 & 0.9164 & 0.91723 & \textbf{0.9660} & \textit{0.9529}\\
\hline
\textbf{Mean AUC} & 0.7451 & 0.841 & 0.85433 & \textit{0.8681} & \textbf{0.8695}\\ \bottomrule
\end{tabular}}
\vspace*{8pt}
\centering
\\
\footnotesize \textbf{Bold} indicates the highest value, and \textit{Italic} indicates the second-highest value.
\end{table}

\begin{figure}
    \centering
    \begin{subfigure}{0.49\linewidth}
        \centering
        \includegraphics[width=\linewidth]{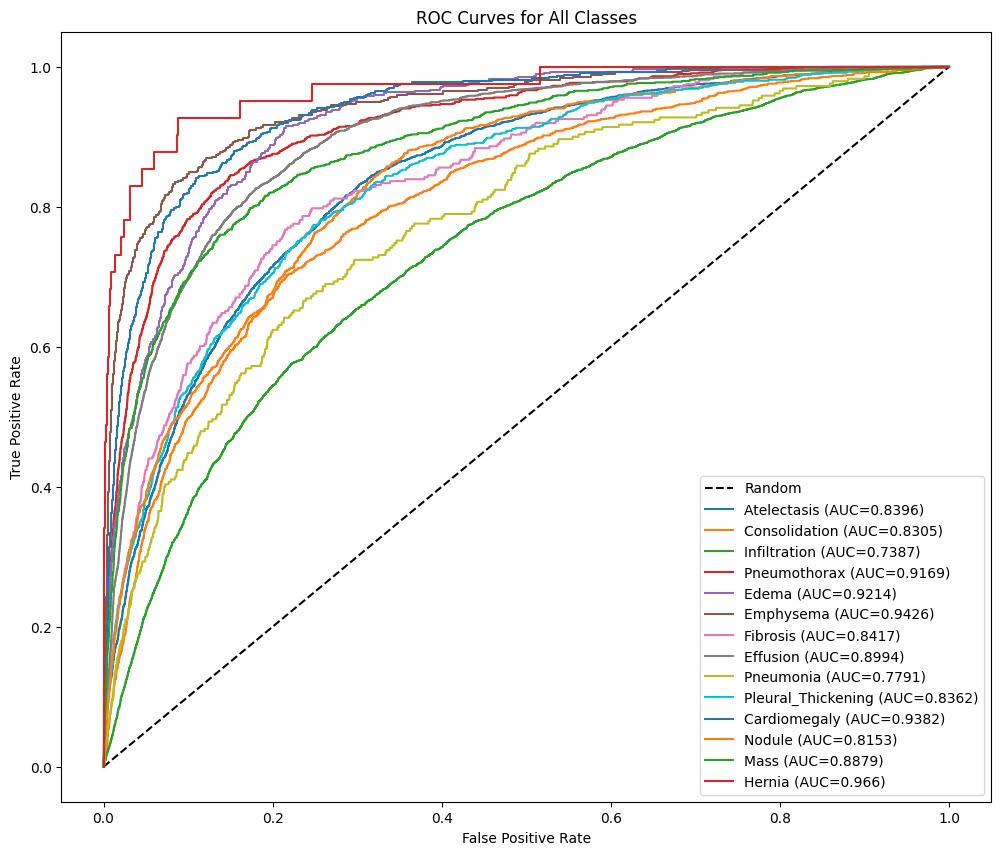}
        \caption{CBAM-Enhanced DenseNet121-based Feature Extractor.}
    \end{subfigure}
    \hfill
    \begin{subfigure}{0.49\linewidth}
        \centering
        \includegraphics[width=\linewidth]{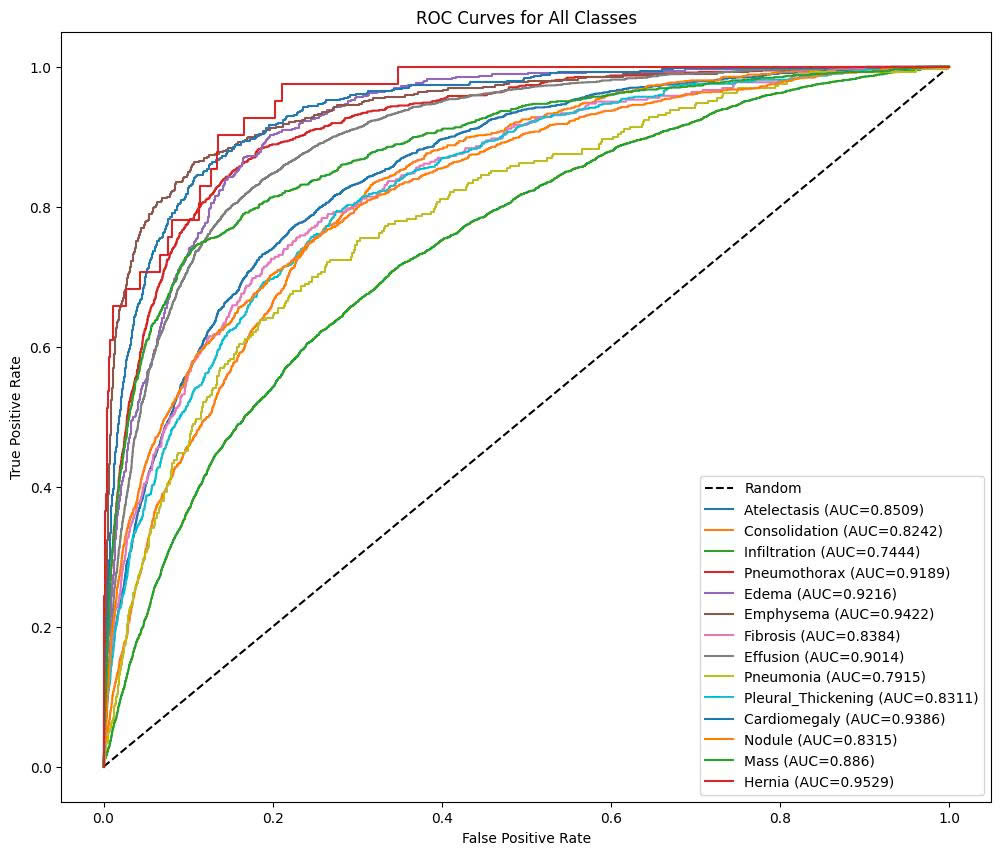}
        \caption{CBAM-Enhanced VGG16-based Feature Extractor.}
    \end{subfigure}
    \caption{ROC AUC of Each Pathology in our Methods.}
    \label{roc_auc}
\end{figure}

The ROC AUC scores for each individual pathology across both models are visualized in Fig.~\ref{roc_auc}, providing a clear comparison of performance between the CBAM-Enhanced DenseNet121 and VGG16-based feature extractors.

\subsection{Ablation Studies}

\subsubsection{Impact of CBAM-Enhanced CNN Blocks Placement on Different Backbone}

\begin{table}
\caption{Impact of CBAM-Enhanced CNN Blocks Placement on Different Backbone.\label{position}}
\resizebox{0.4\textwidth}{!}{{\begin{tabular}{@{}ccc@{}} \toprule
Backbone & Position & Mean AUC \\ \midrule
DenseNet121 & After Dense Block 1, 2, 3, 4 & \textit{0.8543} \\
DenseNet121 & After Dense Block 3, 4 & \textbf{0.8681} \\
DenseNet121 & After Dense Block 4 & 0.8452 \\
\hline
VGG16 & After VGG Block 1, 2, 3, 4, 5 & \textit{0.8689} \\
VGG16 & After VGG Block 3, 4, 5 & \textbf{0.8695}\\
VGG16 & After VGG Block 5 & 0.8218\\
\hline
\end{tabular}}}
\vspace*{8pt}
\centering
\\
\textbf{Bold} - the highest value, \textit{Italic} - the second-highest value.
\end{table}

As presented in Table~\ref{position}, the placement strategy of Convolutional Block Attention Modules (CBAM) substantially influences classification performance for both DenseNet121 and VGG16 backbones. In particular, integrating CBAM into deeper semantic layers, while preserving the integrity of early convolutional stages, yields the most favorable outcomes.

For DenseNet121, the configuration with CBAM inserted after the third and fourth dense blocks achieves the highest mean AUC of \textbf{0.8681}, outperforming both the full-placement variant (blocks 1–4) with an AUC of \textbf{0.8543}, and the minimal configuration (block 4 only) with an AUC of \textbf{0.8452}. These results suggest that attention mechanisms are most effective when applied to later stages where abstract semantic features dominate, while preserving low-level representations in earlier layers.

A similar trend is observed with the VGG16 backbone. The highest mean AUC of \textbf{0.8695} is achieved when CBAM modules are applied after blocks 3, 4, and 5, slightly exceeding the full-placement configuration (blocks 1–5), which attains an AUC of \textbf{0.8689}. In contrast, restricting attention to only the final block (block 5) leads to a marked performance drop (AUC = \textbf{0.8218}), indicating that insufficient attention coverage limits the model’s ability to capture intermediate semantic information.

These observations highlight the critical role of selective attention integration. While CBAM enhances discriminative representation, its effectiveness hinges on judicious placement—balancing abstraction with spatial precision. Both excessive and insufficient usage can impair performance, whereas targeted deployment consistently delivers superior results across architectures.

\subsubsection{Investigate on Training Strategy}

\begin{table}
\centering
\caption{Investigate Training Strategy on Mean AUC.\label{training_strategy}}
\resizebox{\textwidth}{!}{{\begin{tabular}{@{}cccc|ccc@{}} \toprule
Backbone & Position & 1 Stage Training & 1 Stage Training & 2 Stage Training  & 2 Stage Training & 2 Stage Training \\
&& BCE Loss & Focal Loss & BCE Loss & Focal Loss & BCE + Focal Loss \\ \midrule
DenseNet121 & After Dense Block 1, 2, 3, 4 & 0.8328 &0.8360 &0.8543 &0.8547 & 0.8551\\
DenseNet121 & After Dense Block 3, 4 & 0.8384 & 0.8400& \textbf{0.8681} &0.8476 &0.8626\\
DenseNet121 & After Dense Block 4 &0.8218 & 0.8283&0.8452 &0.8459 &0.8445 \\
\hline
VGG16 & After VGG Block 1, 2, 3, 4, 5 &0.8282 & 0.8244&0.8689 & 0.8306&0.8688 \\
VGG16 & After VGG Block 3, 4, 5 &0.8223 & 0.8133&\textbf{0.8695} &0.8433 & 0.8695\\
VGG16 & After VGG Block 5 &0.8131 & 0.8204&0.8218 &0.8394 & 0.8205\\
\hline
\end{tabular}}}
\end{table}

Table~\ref{training_strategy} provides a comprehensive comparison of training strategies, evaluating both single-stage and two-stage paradigms across various loss functions and attention placements. The results consistently favor the two-stage training approach over its single-stage counterparts for both DenseNet121 and VGG16 backbones.

In the two-stage framework, models are initially trained exclusively on abnormal samples to capture class-specific discriminative patterns, followed by fine-tuning on the entire dataset. This sequential strategy helps mitigate overfitting and reduces the dominance of the “No Finding” class, which often introduces distributional bias. Notably, this approach yields the highest mean AUCs across most configurations. For instance, the DenseNet121 variant with CBAM modules inserted after blocks 3 and 4 achieves a peak AUC of \textbf{0.8681} using two-stage Binary Cross Entropy (BCE) loss, substantially surpassing the best single-stage outcome of \textbf{0.8400} attained with Focal Loss. A similar trend is observed for the VGG16 backbone, where the two-stage setup with CBAM at blocks 3, 4, and 5 reaches a top AUC of \textbf{0.8695}.

Interestingly, even when applying imbalance-aware loss functions such as Focal Loss, single-stage training fails to match the performance of the two-stage paradigm. This observation suggests that isolating abnormal data in the initial stage enables the model to more effectively learn inter-class semantics, which are otherwise obfuscated in the full training set due to label imbalance and class dominance. The subsequent fine-tuning phase reinforces global representation learning while curbing overfitting and suppressing negative class bias.

Collectively, these findings not only corroborate the efficacy of curriculum-aware strategies as proposed in~\cite{7}, but also extend them by demonstrating the superiority of abnormal-first training even in the presence of imbalance-handling loss functions. These insights underscore the critical importance of training curricula that are tailored to the unique characteristics of medical imaging datasets, particularly those with long-tailed distributions and noisy annotations concentrated in majority negative classes.

\section{Conclusion}

This paper introduces a Convolutional Block Attention Module (CBAM)-enhanced convolutional neural network (CNN) framework tailored for multi-label classification of chest X-ray images. By embedding CBAM into conventional CNN backbones, the proposed architecture facilitates hierarchical attention mechanisms that selectively amplify anatomically salient features—without the need for pixel-level annotations or auxiliary supervision. 

Empirical evaluations on the ChestXray14 dataset~\cite{1} demonstrate the effectiveness of this design: our framework achieves a mean AUC of \textbf{0.8681} when integrated with DenseNet121, and a slightly higher AUC of \textbf{0.8695} when integrated with VGG16, surpassing previously established baselines.

Future directions include enhancing the clinical interpretability of the attention responses, as well as modeling inter-pathology dependencies to further improve multi-disease diagnostic accuracy and reasoning.

\end{document}